\documentclass{article}
\usepackage{spconf,amsmath,graphicx}
\usepackage{multirow}
\usepackage{multicol}
\usepackage{booktabs}
\usepackage{xcolor}
\usepackage{pifont}
\usepackage{amsfonts}
\usepackage{hyperref}
\usepackage{cleveref}

\newcommand{\cmark}{\textcolor{green}{\ding{51}}}
\newcommand{\xmark}{\textcolor{red}{\ding{55}}}

\title{Establishing Robust Retinal Eye Tracking: A Weakly Supervised Algorithmic Framework}
\name{
\parbox{\textwidth}{\centering
Bo Wen$^{1,2}$, Dillon Lohr$^{1}$, Yatong An$^{1}$, Pushkar Anand$^{1}$, Alexander Fix$^{3}$, Ruobing Qian$^{1}$, Catherine A. Fromm$^{1}$, Yimin Ding$^{1}$, Truong Nguyen$^{2}$, Mohamed El-Haddad$^{1}$, Francesco La Rocca$^{1}$
}
}
\address{$^{1}$Meta Reality Labs, $^{2}$University of California, San Diego, $^{3}$Independent Researcher}

\begin{document}
%
\maketitle
\begin{abstract}
Retinal image-based eye tracking is widely used in ophthalmic imaging and vision science, and is a promising path to deliver higher gaze accuracy than the pupil- and cornea-based approaches commonly used in modern AR/VR devices. Nevertheless, existing retinal tracking algorithms still primarily rely on classical  template-matching registration, which can be insufficiently robust to retinal feature variability and real-world imaging conditions. In this work, we propose a novel weakly-supervised, learning-based framework for robust retinal eye tracking. Initial studies demonstrate high accuracy, achieving the 95th-percentile gaze error $< 0.45^\circ$ across a cohort of 6 participants.

\end{abstract}
\vspace{-0.2cm}

\vspace{-0.1cm}
\section{Introduction}
\label{sec:intro}
\vspace{-0.2cm}

Retinal image-based eye tracking has the potential to deliver substantially higher gaze accuracy than traditional pupil- or cornea-based approaches. This is because it measures gaze more directly, by observing where light falls on the retina—particularly relative to the fovea, which defines the center of vision. The core idea is that each gaze direction corresponds to a uniquely imaged retinal region. By tracking retinal image position relative to the fovea, gaze can be estimated from the resulting image translation. Existing methods \cite{Liu2024, Bedggood2017, Stevenson2016, Sheehy2015} for retinal eye tracking still heavily rely on classical template-matching-based image registration techniques, such as cross-correlation, to estimate frame-to-frame image displacement. However, their performance is not always stable, particularly when large changes in gaze and accommodation induce complex variations in visible retinal image features due to changes in viewing angle and focal depth. 

In addition, state-of-the-art registration methods for conventional retinal images \cite{Registration-SuperJunction, Registration-Wang-TIP, Registration-Zhang-TIP, SuperRetina, Registration-Zhang-ICIP} rely heavily on retinal vessel features (arteries and veins). However, due to system design constraints, many retinal eye tracking systems \cite{MEMSLO, Sheehy2012tslo, Sheehy2015}, especially those targeting high resolution, have a small Field of View (FoV) (typically $< 10^\circ$). As a result, images often lack robust features because vessels are frequently absent or sparse. The remaining non-vessel cues, such as texture arising from the photoreceptor, retinal nerve fiber layer, and other fine-scale retinal microstructure, are substantially less reliable under changes in viewing angle and user alignment. These factors motivate the need for a specialized algorithm that remains robust under such challenging conditions.

To address these challenges, we present a retinal eye tracking method that operates reliably with small-FoV retinal imaging. Our approach is weakly supervised and requires only minimal coarse annotations of a small set of matched points between pairs of overlapping retinal images. The proposed method has two main components. First, multiple images are stitched to construct a larger reference retina map spanning the retinal region used for gaze estimation. In contrast to prior methods which register, stitch and blend multiple images into an explicit reference map (with image degradation due to blending), we introduce a 'canonical feature space' approach that preserves features from the original images while mapping them into a canonical coordinate system. This is crucial since the degradation is difficult to avoid—even with perfect registration—because retinal appearance at the same anatomical location can vary across images even with slight viewpoint and user alignment changes. Registration is then performed implicitly by matching features in this space, from which gaze direction can be derived. Furthermore, we introduce a specialized image registration model with an innovative joint image enhancement and keypoint description, enabling robust feature matching and registration. After a single training procedure, the model supports both image-to-image registration for canonical feature space construction and image-to-feature matching for gaze estimation, resulting in a leaner and more scalable algorithm. The main contributions of this paper are:
\vspace{-0.3cm}
\begin{itemize}
    \item We propose a novel algorithmic framework for retinal eye tracking that significantly improves robustness and accuracy over state-of-the-art methods.
    \vspace{-0.3cm}
    \item We introduce a canonical feature space registration method that enables more robust feature matching and gaze estimation.
    \vspace{-0.3cm}
    \item We present a specialized registration model, featuring a joint image enhancement and feature description approach to address challenging tracking conditions.
    
\end{itemize}

\vspace{-0.4cm}

\section{Related Work}
\label{sec:related_work}
\vspace{-0.2cm}

\subsection{Retinal Image-based Eye Tracking}
\label{ssec:rw_retinal_tracking}
\vspace{-0.2cm}
With the advent of scanning laser ophthalmoscopy (SLO) and adaptive optics SLO (AOSLO), strip-based cross-correlation has become the primary algorithmic paradigm for retinal eye tracking. In this technique, narrow image strips are cross correlated against a reference retinal image to estimate eye motion. Building on earlier use of strip-based cross-correlation in AOSLO systems, the Tracking SLO (TSLO)  demonstrated that adaptive optics is not strictly required to achieve tracking performance comparable to AOSLO-based approaches \cite{Sheehy2012tslo}. Subsequent work extended this paradigm to enable active tracking in closed-loop AOSLO systems \cite{Sheehy2015} and to support binocular retinal tracking \cite{Stevenson2016}, and it remains the foundation of most real-time retinal eye tracking systems.

Some works have proposed accuracy-oriented refinements to strip-based tracking. Bedggood \emph{et al.} introduced distortion-aware retinal eye tracking by jointly estimating eye motion and reference deformation \cite{Bedggood2017}. More recently, Liu \emph{et al.} proposed a substrip-based image montaging algorithm for AOSLO images \cite{Liu2024}. All of these methods retain correlation-based strip matching as the core motion estimation mechanism, which can limit both robustness and accuracy.  

\vspace{-0.4cm}
\subsection{Retinal Image Registration}
\label{ssec:rw_retinal_registration}
\vspace{-0.2cm}
Aligning retinal images acquired from different camera poses and imaging modalities is a fundamental task in retinal image processing. In recent years, deep learning-based methods have come to dominate the retinal image registration literature. For single-modality retinal image registration (e.g. fundus camera imaging), Liu \textit{et al.}~\cite{SuperRetina} proposed a semi-supervised feature detection and description approach that iteratively refines ground truth keypoint locations during training. Wang \textit{et al.}~\cite{Registration-SuperJunction} introduced a joint vessel segmentation and keypoint detection framework that constrains keypoints to vessel junctions to improve robustness. Sivaraman \textit{et al.}~\cite{SIVARAMAN2025} proposed combining the use of a SIFT keypoint detector \cite{Lowe1999SIFT} with a DIFT descriptor \cite{tang2023DIFT} to establish correspondences between paired retinal images for registration. For cross-modality retinal image registration, Wang \textit{et al.}~\cite{Registration-Wang-TIP} proposed extracting vessel maps and then detecting and matching keypoints across modalities to estimate the registration transform. Numerous post-processing methods have also been proposed for pixel-level refinement. For example, Zhang \textit{et al.}~\cite{Registration-Zhang-TIP} estimated optical flow between coarsely registered retinal images to improve alignment and, in subsequent work, leveraged a 3D eyeball model to correct distortion in ultra-wide-field retinal images for more precise registration \cite{Registration-Zhang-ICIP}. However, most of these methods rely heavily on the retinal vessel features, which are not always visible in high resolution, small FoV retinal images, particularly when acquired near the foveal region. This highlights the need for a specialized registration method for retinal tracking images, which can still operate robustly under limited vessel features.

\vspace{-0.3cm}
\section{Proposed Method}
\label{sec:method}
\vspace{-0.3cm}
\subsection{Overview}
\vspace{-0.2cm}
\label{ssec:algorithm overview}
The core principle of retinal eye tracking is that gaze direction can be inferred from how retinal features shift in the captured image as the eye rotates. Specifically, the gaze (pitch, yaw) is related to the translation of a source retinal image relative to a reference (foveal) retinal image acquired when the user looks straight ahead, which we define as the \((0^{\circ}, 0^{\circ})\) gaze direction. This follows from the geometry of retinal imaging and is analogous to panoramic imaging: changes in viewing angle induce predictable image-plane translations, described in general by trigonometric projection, and, in the small-angle regime, by a linear relationship between image translation and gaze angle. Accordingly, retinal eye tracking can be formulated as estimating the translation of each source image with respect to the reference image, yielding a unique mapping between retinal image location and gaze direction. However, because each image covers only a limited retinal area, at larger gaze angles there may be little-to-no overlap between the \((0^{\circ}, 0^{\circ})\) gaze reference image and the source image. Consequently, we first construct a wider FoV `reference map' (with the \((0^{\circ}, 0^{\circ})\) gaze image anchored at the center), and then register each source image to this map to estimate its location relative to the center.

\begin{figure*}[ht]
\centering
\includegraphics[width=0.99\textwidth]{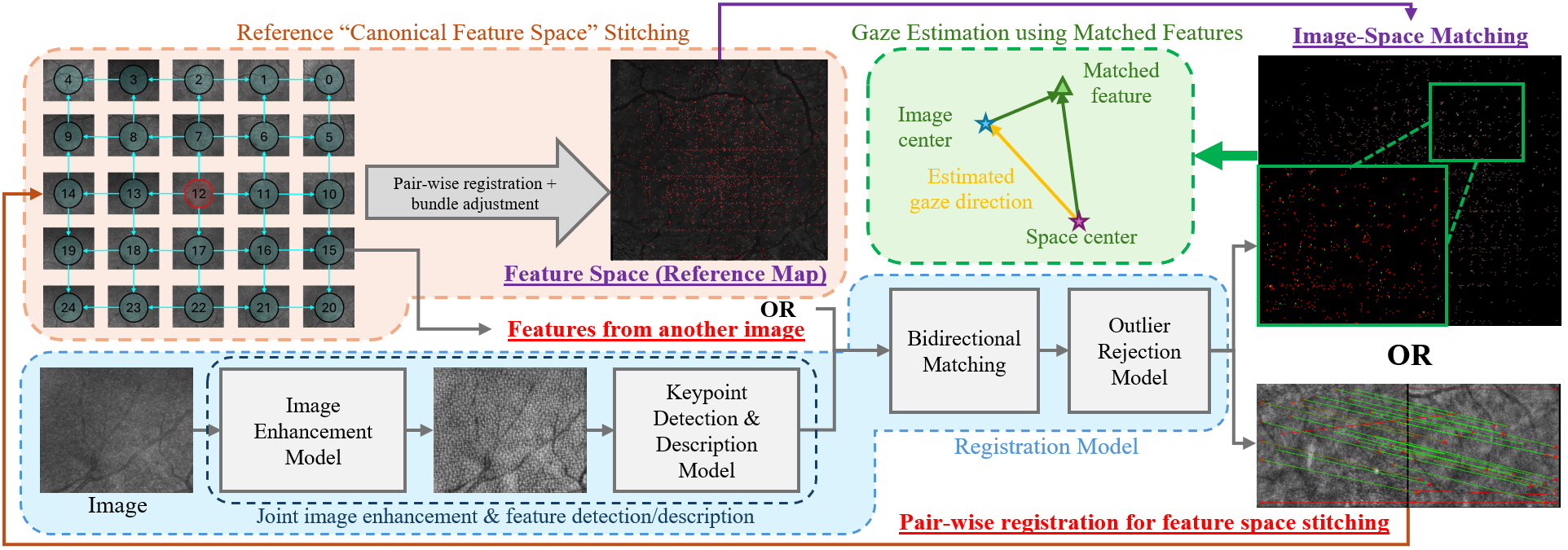}
\vspace{-0.5cm}
\caption{Overview of the proposed method. (Best viewed in color; zoom in for details.)}
\label{fig:overview}
\vspace{-0.5cm}
\end{figure*}

To mitigate image blending artifacts and feature degradation during reference map stitching, we propose stitching a `canonical feature space'. As illustrated in Fig. \ref{fig:overview}, we first acquire a rectangular grid scan of the retina. Starting from the central \((0^{\circ}, 0^{\circ})\) gaze image (red node 12), we perform pairwise registration between adjacent images and apply bundle adjustment to estimate the locations of all images within the scan. We then retain the detected features (keypoints) at their estimated locations and transform them into a shared canonical coordinate system whose origin is defined by the center of the \((0^{\circ}, 0^{\circ})\) gaze image. Subsequently, we establish feature correspondences between each source image and the canonical feature space, then estimate the gaze for each source image by (i) locating the image center in canonical coordinates, and (ii) converting the resulting pixel location to gaze angle via system calibration.

\begin{figure}[ht]
\centering
\includegraphics[width=0.47\textwidth]{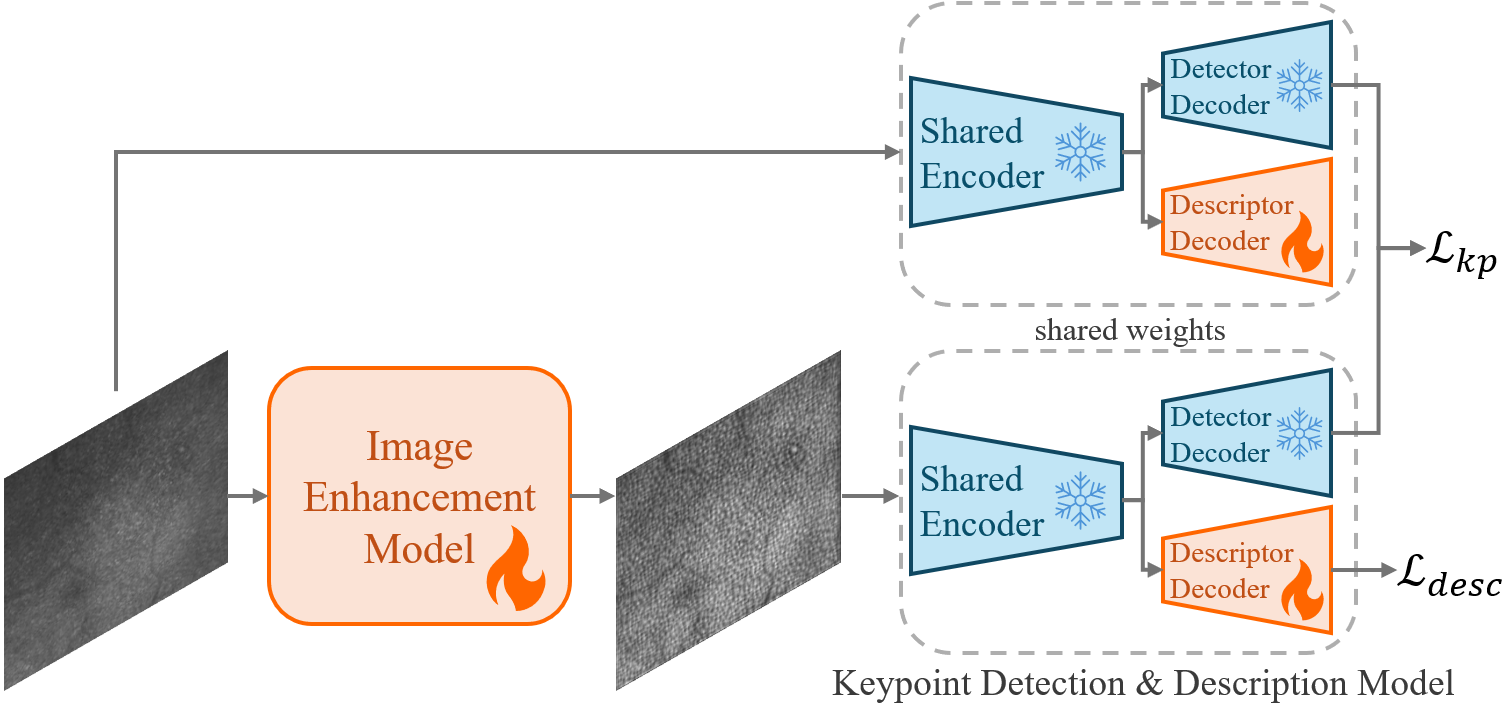}
\vspace{-0.4cm}
\caption{Training pipeline for the proposed joint image enhancement and keypoint detection/description model.}
\label{fig:enhance_kp}
\vspace{-0.6cm}
\end{figure}

\vspace{-0.4cm}
\subsection{Weakly Supervised Registration Model}
\label{ssec:registration_model}
\vspace{-0.2cm}

The proposed algorithm comprises two stages of feature detection, matching and registration. In the first stage, we construct the canonical feature space by registering paired retinal images. In the second stage, we detect features in each source image and match them to the canonical feature space to estimate gaze. Because the visibility of retinal features can change substantially—particularly under large changes in gaze and accommodation—we introduce a specialized registration model that robustly supports both stages.

As shown in Fig. \ref{fig:overview}, we first propose a joint image enhancement and keypoint description method (Fig. \ref{fig:enhance_kp}). Our approach uses a lightweight image enhancement network (Zero-DCE) \cite{zero-dce} as the backbone to estimate per-pixel polynomial curve adjustment for the input image:
\vspace{-0.2cm}
\begin{equation}
E(I;\alpha) = I+\alpha I(1-I)
\end{equation}
where $E(\cdot)$ denotes the enhanced image at the current iteration, $I$ is the image from the previous enhancement iteration and $\alpha$ is the predicted per-pixel curve adjustment parameter. We apply the enhancement for N=2 iterations, initializing $I$ in the first iteration with the raw image. During training, both the enhanced and raw images are passed to the keypoint detection and description network. We adopt SuperPoint \cite{SuperPoint} as the backbone, which consists of a shared encoder with two decoder heads: a keypoint detector and a descriptor. Because ground truth keypoints are infeasible to obtain, we use weights pretrained on a large synthetic, non-retinal dataset \cite{SuperPoint} and freeze the shared encoder and detector decoder. We then fine-tune the descriptor decoder using a triplet loss:
\vspace{-0.2cm}
\begin{equation}
\mathcal{L}_{desc}=\sum_{i\in\mathcal{K}}^{}max(0, m+\phi_{pos} - \frac{1}{2}(\phi_{neg-rand}+\phi_{neg-hard}))
\end{equation}
We refer readers to \cite{SuperRetina} for details of this loss. Furthermore, we propose a keypoint-preserving and boosting loss: 
\vspace{-0.3cm}
\begin{equation}
\begin{aligned}
\mathcal{L}_{kp} &=max(0, h-[\sum_{i\in\mathcal{P}}^{}\sigma(\frac{D_{enhanced}^{i} - \gamma}{t}) \\
 &-\textbf{stopgrad}(\sum_{i\in\mathcal{P}}^{}\sigma(\frac{D_{raw}^{i} - \gamma}{t}))])
\end{aligned}
\end{equation}
where $D_{enhanced}^{}$ and $D_{raw}^{}$ denote the keypoint probability maps produced by the pretrained, frozen detector decoder, $\sigma$ is the sigmoid function, $\gamma$=0.1 is the threshold used to determine whether a pixel corresponds to a keypoint, $t$=0.1 is a temperature hyperparameter, and $h$=1000 is the number of additional keypoints that we encourage to be detectable in the enhanced image by the pretrained detector. The loss is summed over all pixels $\mathcal{P}$ in the keypoint probability map. By jointly optimizing with the descriptor loss $\mathcal{L}_{desc}$, the enhancement network learns to produce images in which the pretrained detector identifies more keypoints (which empirically improves registration performance) and, for each detected keypoint, yields more robust descriptors.

After extracting keypoints and their corresponding descriptors from a retinal tracking image, we establish correspondences either with features from another image or with features from the canonical feature space. We first compute a set of coarse correspondences via bi-directional (mutual nearest-neighbor) descriptor matching: two keypoints $m$ and $n$ from the two images are matched if and only if each is the nearest neighbor of the other in descriptor space. These coarse correspendences are then passed to an outlier rejection network, which predicts a confidence score in $\in [0, 1]$ for each candidate match. We use the attentional graph neural network from SuperGlue \cite{SuperGlue} as the backbone, adapted to our weakly supervised settings in which dense ground truth correspondences are unavailable and only an approximate ground-truth translation matrix is provided for each image pair. Since the inputs are pre-matched keypoint pairs, the model outputs a square score matrix of size $N \times N$, where $N$ is the number of coarse correspondences. The diagonal entries correspond to the confidence scores of the $N$ input correspondences. We train the model as a binary classification problem, predicting whether each correspondence is an inlier (1) or outlier (0), using the binary cross-entropy loss:
\vspace{-0.3cm}
\begin{equation}
\mathcal{L}_{cls}=-\frac{1}{N}\sum_{i}^{N}y_i\log(x_i)+(1-y_i)\log(1-x_i)
\end{equation}
where $x$ denotes the predicted score for correspondence $i$, and 
\begin{equation}
y=\left\{\begin{matrix}
 1, \ ||\mathcal{T}(k_{src}, M_{gt})-k_{tgt}||_{2} < \epsilon \\
 0, \ otherwise \\
\end{matrix}\right.
\end{equation}
where $k_{src}$ denotes the keypoint from the first image in a correspondence, $k_{tgt}$ denotes its matched keypoint in the second image, and $M_{gt}$ is the coarse ground truth translation matrix. We set $\epsilon$=10 pixels as the threshold for labeling a correspondence as positive. To improve robustness when matching to the canonical feature space, we apply a random bias to the target-image keypoint locations during training. This augmentation encourages the model to generalize beyond simply rejecting outliers between paired raw images. Training of the registration model relies solely on the coarse ground truth translation matrix $M_{gt}$, which is obtained by manually annotating 3+ correspondences for each training image pair.

\vspace{-0.2cm}
\subsection{Canonical Feature Space Generation}
\label{ssec:canonical feature space}
As shown in Fig. \ref{fig:overview} and described in Section \ref{ssec:algorithm overview}, after performing pairwise registration between adjacent images, we apply bundle adjustment to estimate the spatial pose of each image by solving:
\begin{equation}
\min_{n}||W^{\frac{1}{2}}(C\mathbf{n}-\mathbf{\mu})||^{2}
\end{equation}
where, for a grid graph such as in Fig. \ref{fig:overview} with M edges and N nodes, $\mathbf{n}\in \mathbb{R}^{N\times2}$ denotes the unknown 2D node locations (image positions) to be estimated, and $\mathbf{\mu}\in\mathbb{R}^{M\times2}$ denotes the pairwise translations estimated by the registration model for each edge. $C\in \mathbb{R}^{M\times N}$ is the oriented node-edge incidence matrix, with one row per edge. For an edge directed from node A to node B, the corresponding row has -1 in the column for node A, +1 in the column for node B, and zeros elsewhere. $W=\mathbf{diag}([X_1, X_2, ... , X_M])\in \mathbb{R}^{M\times M}$ is a diagonal weight matrix, where $X_i$ is the sum of correspondence confidence scores $x$ associated with edge $i$. After estimating the node locations $\mathbf{n}$, we express all keypoint coordinates in a shared canonical coordinate system with its origin at the center pixel of the central image, while leaving the descriptors unchanged. We keep all features matched during pairwise registration in this canonical feature space.

\begin{table*}[ht]
    \caption{Comparison of gaze estimation errors between the proposed method and prior registration methods.}
    \label{tab:main_comp}
    \centering

    \begin{tabular}{c|cccc|cccc}
        \toprule
        \multirow{2}{*}{Method} & \multicolumn{4}{c|}{Fake Eye} & \multicolumn{4}{c}{Real Eyes} \\
        \cmidrule{2-9}
        & Mean(std) $\downarrow$ & E95$\downarrow$ & E75$\downarrow$ & E50$\downarrow$ & Mean(std)$\downarrow$ & E95$\downarrow$ & E75$\downarrow$ & E50$\downarrow$ \\
        
        \midrule

        SIFT (Lowe et al., 1999) \cite{Lowe1999SIFT}&0.27(±0.21)°&0.60°&0.48°&0.26°&4.34(±1.62)°&7.16°&5.31°&4.28°\\
        ORB (Rublee et al., 2011) \cite{Rublee2011ORB}&5.46(±1.38)°&7.74°&6.30°&5.56°&5.17(±1.45)°&7.61°&5.92°&5.19°\\
        Superpoint (DeTone et al., 2018) \cite{SuperPoint}&0.28(±0.23)°&0.64°&0.50°&0.29°&1.88(±1.86)°&5.52°&3.20°& 1.20°\\
        Superglue (Sarlin et al., 2020) \cite{SuperGlue}&0.56(±0.81)°&1.74°&0.52°&0.32°&1.41(±1.08)°&3.55°&1.85°&1.20°\\
        Substrip NCC (Liu et al., 2024) \cite{Liu2024}&0.24(±0.18)°&0.54°&0.41°&0.24°&2.10(±2.77)°&8.06°&3.38°&0.32°\\
        SIFT+DIFT (Sivaraman et al., 2025) \cite{SIVARAMAN2025}&3.02(±1.48)°&5.80°&3.96°&2.83°&3.50(±1.71)°&6.55°&4.62°&3.30°\\
        \textbf{Our Proposed} &\textbf{0.09(±0.07)°}&\textbf{0.23°}&\textbf{0.13°}&\textbf{0.07°}&\textbf{0.18(±0.16)°}&\textbf{0.38°}&\textbf{0.23°}&\textbf{0.16°}\\

        \bottomrule
    \end{tabular}
    \vspace{-0.5cm}
\end{table*}

\begin{table}[ht]
    \vspace{-0.4cm}
    \caption{Comparison of experimental settings across trials in the real-eye dataset. BC denotes background color.}
    \label{tab:trial_div}
    \centering
    \vspace{+0.2cm}

    \begin{tabular}{c|ccc}
        \toprule
        Trial & Gaze Target & Pupil Steering & BC(pupil size) \\
        \midrule
        0& rectangle grid &\xmark & black(large)\\
        1& rectangle grid &\cmark & black(large)\\
        2& random &\cmark & black(large)\\
        3& random &\cmark & gray(medium)\\
        4& random &\cmark & white(small)\\

        \bottomrule
    \end{tabular}
    \vspace{-0.5cm}
\end{table}

\begin{table}[ht]
    \caption{Ablation study of the proposed components and their contributions to gaze estimation accuracy on the real-eye dataset. JIE denotes joint image enhancement. KDD denotes keypoint detection and description (used in all experiments as a necessary step). OR denotes outlier rejection. CFSR denotes canonical feature space registration. Without CFSR means registration is performed against an explicit reference retinal map generated using the estimated translation, with images fused via a multi-band blending \cite{mbblending1983}. }
    \label{tab:ablation}
    \centering

    \begin{tabular}{cccc|cc}
        \toprule
        JIE & KDD & OR & CFSR & Mean(std) $\downarrow$ & E95$\downarrow$ \\
        \midrule
        \xmark&\cmark&\cmark&\cmark&0.247(±0.335)°&0.554°\\
        \cmark&\cmark&\xmark&\cmark&0.694(±0.380)°&1.405°\\
        \cmark&\cmark&\cmark&\xmark&4.507(±3.206)°&10.179°\\
        \cmark&\cmark&\cmark&\cmark&\textbf{0.182(±0.156)°}&\textbf{0.379°}\\

        \bottomrule
    \end{tabular}
    \vspace{-0.6cm}
\end{table}

\vspace{-0.3cm}
\subsection{Final Gaze Estimation}
\label{ssec:canonical feature space}
For each source image in a test video sequence, we apply the same enhancement and keypoint detection/description model to extract features. We then match these features to the canonical feature space using bi-directional matching followed by the outlier rejection model. For each retained correspondence $(k_{src}, k_{space})$, we define the gaze displacement as the vector from the canonical feature space center to the source image center, computed as $\mathbf{t}_{space}-\mathbf{t}_{src}$, where $\mathbf{t}_{space}$ is the location of $k_{space}$ in the canonical feature space coordinate system and $\mathbf{t}_{src}$ is the location of $k_{src}$ in the source image coordinate system (with the origin at the image center). We convert this displacement from pixels to degrees using the pixel-to-degree ratio estimated during system calibration. The final gaze estimate for the image is computed as the correspondence-score-weighted average over all retained matches, where the weights are the confidence scores predicted by the outlier rejection model.

\vspace{-0.4cm}
\section{Experiment}
\label{sec:experiment}
\vspace{-0.4cm}
\subsection{Dataset}
\label{ssec:dataset}
Experiments were conducted on both phantom-eye and real-eye images over a +/-5$^{\circ}$ gaze range. For the phantom-eye experiments, we used a dataset collected with a custom retinal eye tracking system \cite{MEMSLO}. Ground truth gaze direction (pitch and yaw) was provided by the motorized goniometer stages holding the phantom eye. The dataset contains 25 images for training, 25 for validation and 1457 for testing. For real-eye experiments, we used a dataset acquired with the same imaging system from 6 human volunteers. In this case, ground truth gaze is inferred from the fixation target location, and is therefore subject to small, unavoidable inaccuracies due to involuntary eye movements. To mitigate this issue, for each gaze direction we used only the five frames (0.17s) immediately following a click event, where subjects clicked a mouse once they were confident they were precisely fixating on the target. In total the real-eye dataset includes 205 images for training, 205 for validation, and 4064 for testing. All images have resolution $253\times207$. As shown in Table \ref{tab:trial_div}, the test set comprises five trials designed to evaluate the algorithm under different experimental conditions. These conditions vary in (i) the pattern of gaze-target locations, (ii) whether pupil steering—a hardware mechanism in the retinal tracking system— was employed (without pupil steering, gaze is estimated solely from retinal image displacement; with pupil steering, an additional offset determined by system calibration is applied to the displacement-based estimate \cite{MEMSLO}), and (iii) the monitor background color used to display the target, which affects pupil size and retinal image appearance. The pixel/degree ratios for this system are (40.18, 40.17) in the x (yaw) and y (pitch) directions, respectively.

\vspace{-0.2cm}
\subsection{Implementation Details}
\label{ssec:imple_details}
For the joint image enhancement and keypoint description network, we used a learning rate of 5e-4 and trained for 600 epochs on the phantom-eye dataset and 2000 epochs on the real-eye dataset. For the outlier rejection network, we used a learning rate of 2e-5 and trained for 600 epochs on both datasets. Keypoints were sampled using non-maximum suppression with a 7x7 window and a detection threshold 0.15. For the keypoint preserving loss, we used a margin $h$ of 2000 keypoints and a weight of 5e-5; this term was added to the descriptor loss to form the final loss function. During training, we paired images depicting adjacent retinal regions and constructed an adjacency graph over all training images (per subject). For each randomly selected source image, we then sampled a target image by randomly traversing the graph for one or two steps. All models were implemented in Python using PyTorch.

\begin{figure}[ht]
\centering
\includegraphics[width=0.48\textwidth]{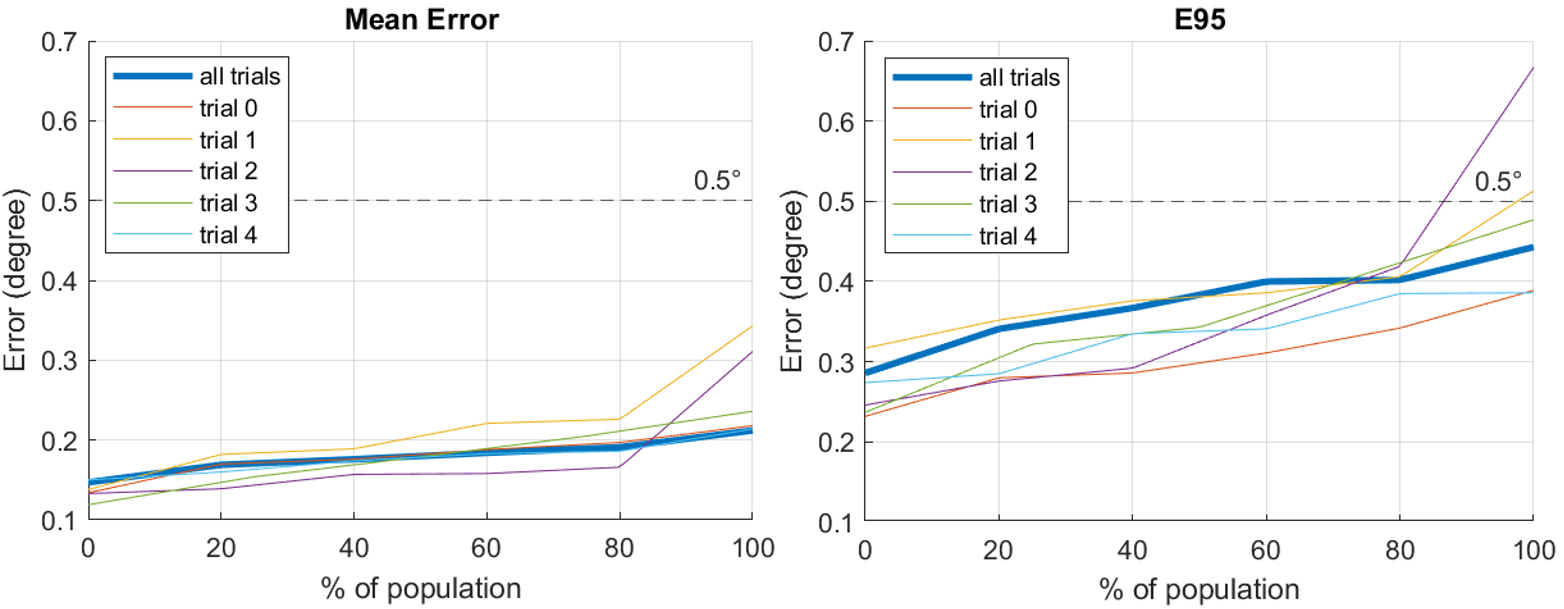}
\vspace{-0.6cm}
\caption{Per-trial population coverage curves on the real-eye dataset.}
\vspace{-0.6cm}
\label{fig:pcc}
\end{figure}

\vspace{-0.3cm}

\subsection{Main Results}
\label{ssec:main_results}
We compared our proposed method with several classical and state-of-the-art general-purpose registration approaches, as well as a state-of-the-art retinal eye-tracking registration method by Liu \textit{et al.}~\cite{Liu2024}, as summarized in Table \ref{tab:main_comp}. All keypoint-based baseline methods use the same bi-directional matching and use RANSAC for outlier rejection except for SuperGlue. In addition to the mean gaze estimation error, we also report the $n$th-percentile error $E_n$ for $n \in \{50, 75, 95\}$. The results show that our method achieves substantially improved accuracy and robustness compared with existing approaches, particularly on real-eye data, where the images are more challenging. Furthermore, Fig. \ref{fig:pcc} shows the population coverage curves of the mean error and $E_{95}$ across trials, highlighting the robustness of the proposed algorithm across different eyes and experimental configurations.

\vspace{-0.3cm}
\subsection{Ablation Study}
\label{ssec:main_results}
\vspace{-0.1cm}
We further report an ablation study in Table \ref{tab:ablation} to quantify the contribution of each  proposed component to tracking performance. The results show that joint image enhancement improves both robustness and accuracy, whereas removing the outlier rejection module leads to a substantial performance drop. Finally, the proposed canonical feature space registration mitigates feature degradation caused by image blending and enables reuse of the same registration model for reference map generation in downstream gaze estimation.

\section{Conclusion and Future Work}
\label{sec:conclusion}
In this paper, we propose a robust, accurate and practical algorithmic framework for retinal image-based eye tracking. The proposed approach includes multiple methodological contributions, including a task-specialized image registration model and a complementary feature space registration strategy designed to improve robustness to retinal appearance variability. Experiments on both a phantom eye and real eyes across a range of experimental configurations demonstrate consistently improved accuracy and robustness relative to prior retinal tracking approaches. Future work may investigate fine-alignment stages (e.g., optical flow-based refinement) to further improve registration precision, and may expand evaluation to larger and more diverse cohorts spanning broader gaze ranges and more challenging imaging conditions.

\bibliographystyle{IEEEbib}
\bibliography{strings,refs}

\section{Supplementary Materials}
\subsection{Runtime Analysis}
The inference time of the proposed algorithm is presented in Table \ref{tab:runtime}, which yields an approximate 14.5 FPS. The experiment is run on one NVIDIA RTX 3080 GPU, with test image size 253$\times$207 and batch size of 1. The canonical feature space is constructed once per subject in approximately 3.8 seconds (on the same GPU), comprising 40 pairwise registrations across the 5×5 grid scan with bundle adjustment adding negligible overhead ($<$6 ms). This one-time cost does not add to per-frame latency. The primary bottleneck in per-frame inference is the outlier rejection network (67\% of total), which could be reduced in future work by adopting a more efficient backbone.

\begin{table}[ht]
    \caption{Inference time of the proposed algorithm breakdown.}
    \vspace{+0.2cm}
    \label{tab:runtime}
    \centering

    \begin{tabular}{c|c}
        \toprule
        Stage & Latency\ (ms) \\
        \midrule
        Image Enhancement\ (JIE) & 6.4 \\
        Keypoint Detection \& Description\ (KDD) & 6.8 \\
        NMS + Descriptor Sampling & 6.0 \\
        Bi-directional Matching & 2.9 \\
        Outlier Rejection (OR) (SuperGlue) & 46.5 \\
        Gaze Computation & 0.3 \\
        \textbf{Total per sample} & 68.9 \\

        \bottomrule
    \end{tabular}
\end{table}

\subsection{Additional Implementation Details}

\subsubsection{Cross-subject Experimental Settings}
We train a single model across subjects rather than per subject. The training/validation set includes data from 4 of the 6 subjects; the remaining 2 are completely unseen during training. The model performs comparably or better on these unseen subjects, demonstrating cross-subject generalization without reliance on subject-specific cues. 

\subsubsection{Training Pairs Annotation Details}
We labeled "3+" pairs of correspondences for each training image pair, i.e., at least 3 corresponding point pairs per image pair (often more, depending on the number of identifiable correspondences, but usually no more than 10 pairs) are labeled to estimate the approximated 2D ground truth translation matrix. Less than 3 pairs can cause large inaccuracy in the ground truth translation matrix. Additionally, the manual labeling is done for only adjacent images (Fig.1 top-left in the main paper) from trial 0 (for each subject used for training, correspondence between selected images from trial 3/4 and their closest image in the central 9 nodes in the 5x5 grid from trial 0 are also labeled and the images from trial 3/4 are set as one node away from the closest trial 0 node). During the training, instead of relying on fixed annotated pairs from only adjacent nodes, a random pair that is $\leq$ 2 nodes away is sampled and used. The accompanying ground truth translation matrix is generated by accumulating the manual per-adjacent node matrix.

\subsubsection{Baseline Implementation}
All baselines share the identical reference map construction pipeline and differ only in registration method. For all baseline registration methods, we follow the recommended settings in their original papers.

\subsection{Canonical Feature Space Generation Robustness Analysis}
An interesting question is how to handle pairwise registration failures in the canonical feature space generation. In this work, we ensure robust space generation via (1) increasing detectable features from challenging imaging conditions by the joint image enhancement; (2) having controllable grid scan for the feature space - we found $\approx$50\% overlap between adjacent images ensures reliable registration for images from our hardware. The overlap, if needed, can be further increased to avoid the failures and (3) bundle adjustment weighted by the correspondence confidence score estimated by the outlier rejection network down-weights unreliable edges in the graph. To quantify this robustness, we first systematically removed one edge (5$\times$5 grid graph, 25 nodes, 40 edges) at a time and re-ran the bundle adjustment: all 40 single-edge removals keep the graph connected, with worst-case node position shift $<$ 0.001°. Furthermore, Table \ref{tab:robustness} presents a simulation study to test robustness to measurement noise by adding Gaussian noise to all edge measurements simultaneously.

\begin{table}[ht]
    \caption{Canonical feature space robustness simulation study. Even in the extreme case with 5-pixel noise on every edge measurement, the worst-case node error is only $\approx$0.3°.}
    \vspace{+0.2cm}
    \label{tab:robustness}
    \centering

    \begin{tabular}{c|c}
        \toprule
        \textbf{Pairwise Noise (std, pixel)} & Max Node Error (deg) \\
        \midrule
        1.0 & 0.061 ± 0.013 \\
        2.0 & 0.123 ± 0.028 \\
        5.0 & 0.309 ± 0.064 \\
        10.0 & 0.612 ± 0.127 \\

        \bottomrule
    \end{tabular}
\end{table}

\end{document}